\RequirePackage{amsthm}

\documentclass[sn-mathphys-num]{sn-jnl}

\usepackage{graphicx}%
\usepackage{multirow}%
\usepackage{amsmath,amssymb,amsfonts}%
\usepackage{amsthm}%
\usepackage[title]{appendix}%
\usepackage{xcolor}%
\usepackage{textcomp}%
\usepackage{manyfoot}%
\usepackage{booktabs}%

\usepackage{algorithmicx}%
\usepackage{algpseudocode}%
\usepackage{listings}%
\usepackage{enumitem}
\usepackage{lmodern}
\usepackage{algorithm}%

\theoremstyle{thmstyleone}%
\theoremstyle{thmstyletwo}%

\theoremstyle{thmstylethree}%

\begin{document}

\title[Article Title]{Research on Image Processing and Vectorization Storage Based on Garage Electronic Maps}


\author[1]{\fnm{Nan} \sur{Dou}}\email{doudoudounan@163.com}

\author[1]{\fnm{Qi} \sur{Shi}}\email{shiqi0524@163.com}

\author[1]{\fnm{Zhigang} \sur{Lian}}\email{lianzg@sdju.edu.cn}

\affil[1]{\orgdiv{School of electronic Information}, \orgname{Shanghai DianJi University}, \orgaddress{\city{Shanghai}, \postcode{201306}, \country{China}}}


\abstract{For the purpose of achieving a more precise definition and data analysis of images, this study conducted a research on vectorization and rasterization storage of electronic maps, focusing on a large underground parking garage map. During the research, image processing, vectorization and rasterization storage were performed. The paper proposed a method for the vectorization classification storage of indoor two-dimensional map raster data. This method involves converting raster data into vector data and classifying elements such as parking spaces, pathways, and obstacles based on their coordinate positions with the grid indexing method, thereby facilitating efficient storage and rapid querying of indoor maps. Additionally, interpolation algorithms were employed to extract vector data and convert it into raster data. Navigation testing was conducted to validate the accuracy and reliability of the map model under this method, providing effective technical support for the digital storage and navigation of garage maps.}

\keywords{electronic maps, vectorization storage, rasterization storage, classification storage, vehicle navigation}



\maketitle

\section{Background of map digital storage}\label{sec1}

This paper conducts research on the vectorization of electronic maps. When maps are directly scanned into the computer, data is stored in raster format. However, for more accurate location, length, and size definitions, as well as for Geographic Information Systems (GIS) data analysis, spatial data topology analysis, and joint analysis of spatial and attribute data, it is necessary to convert raster data into vector data. This process is known as vectorization \cite{2016A}, which transforms image data into graphic data while preserving the corresponding topological structure. Graphic data includes lines, circular arcs, curves, character strokes, etc. Vector data has higher accuracy, allowing for more precise definition of location, length, and size. Therefore, in GIS data analysis, transformation, spatial data topology analysis, and the joint analysis of spatial and attribute data, vector data is indispensable. Thus, converting raster data from scanned images into vector data becomes one of the important tasks in GIS technology \cite{1959A}.

Raster to vector data conversion is a data transformation technique used to convert between raster and vector representations. Each representation has its own advantages, disadvantages, and applicable scenarios, so the decision to perform raster to vector data conversion depends on the intended use \cite{li2007gis}.

\section{Overview of Digital Storage Issues for Garage Maps}\label{sec2}
\subsection{Problem Statement}
In a large underground parking garage with multiple entrances and parking spaces, the system needs to process its structural map through image processing and convert it into vector data for storage, facilitating the system's map modeling \cite{1020348882.nh}.
\subsection{Research Methodology}
\begin{figure}[ht]
	\centering
	\includegraphics[width=0.8\textwidth]{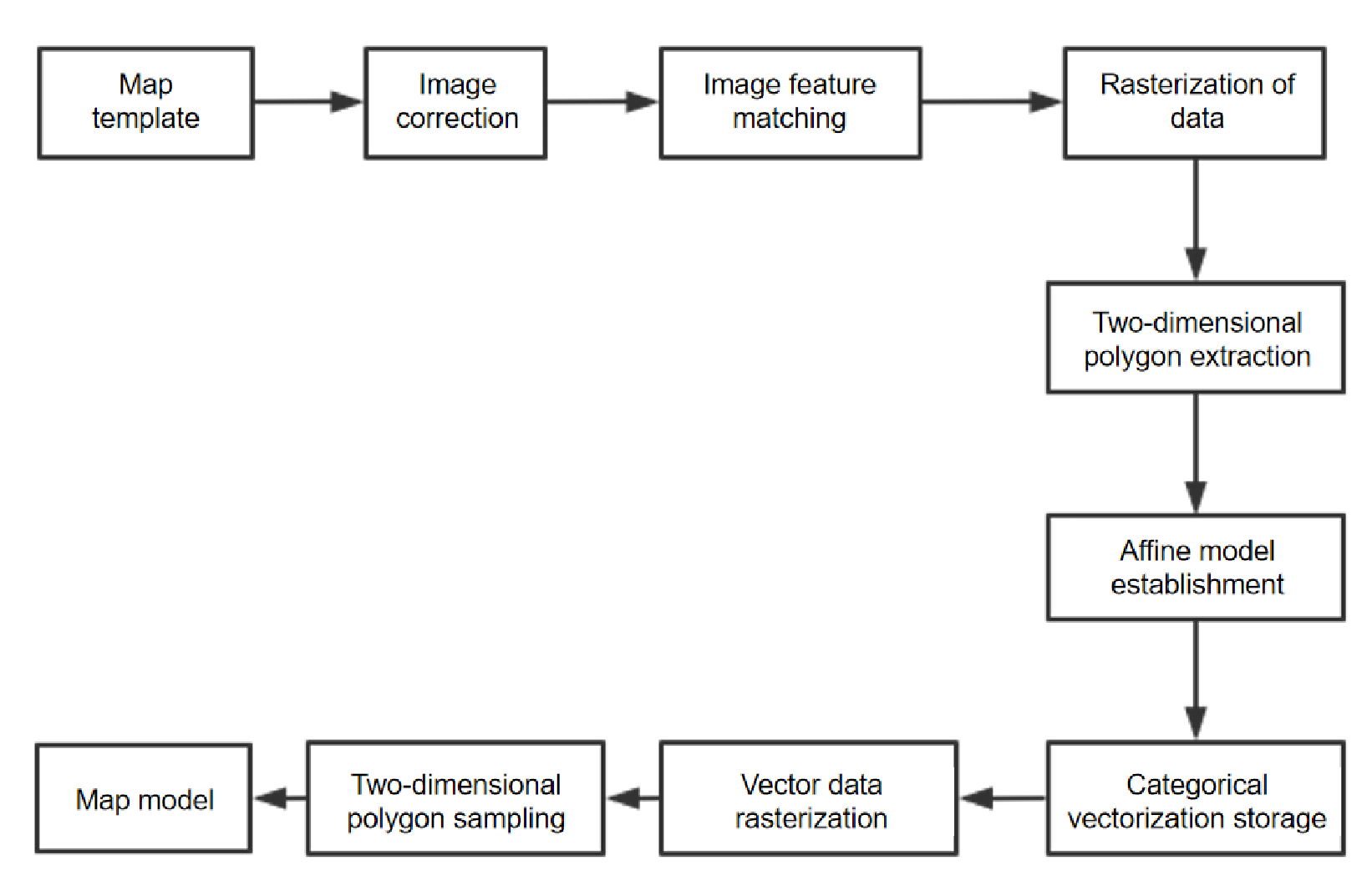}
	\caption{Technical road map}
	\label{fig1}
\end{figure}

This article first uses image feature retrieval to match feature points between the corrected raster image and the latitude and longitude template image. Then, based on the matching results, it obtains the latitude and longitude information and the horizontal and vertical coordinate information on the raster image. Next, it solves for the parameters of the affine transformation model \cite{he2011new}. Finally, it vectorizes the raster image and performs coordinate transformation. The technical approach is illustrated in Figure \ref{fig1} \cite{yang2021rapid}.

The digital storage methods for indoor maps primarily include rasterization, vectorization, hybridization and so on. Among them, rasterization \cite{ali2021performance} is a commonly used method, converts indoor maps into pixel grid data and achieves digital storage through array-based representation. Vectorization transforms indoor maps into vector data, enabling high-precision processing and visual representation. Hybridization combines raster and vector data, comprehensively leveraging their advantages to achieve better results. Indoor maps refer to maps that describe and depict indoor spaces. With the continuous development of artificial intelligence technology and indoor navigation applications, the demand for indoor maps is growing. This paper proposes a method for vectorization and categorical storage of indoor two-dimensional raster data. By converting raster data into vector data and adopting a categorical storage approach, this method achieves efficient storage and rapid querying of indoor maps.

\section{Image Feature Matching and Rasterization Processing}\label{sec3}
\subsection{Image Feature Matching}
For image feature matching \cite{hui2021image}, this paper sequentially matches the template image with the raster image, obtaining the coordinates of the matched blocks. The paper detects contours in the images and calculates the coordinates of the four vertices of the second largest area contour. In longitude matching, the paper calculates the image region from 0 to X\(_ {min}\), while in latitude matching, it calculates the image interval from top to bottom, ranging from 0 to Y\(_{min}\) . X\(_{min}\) is the minimum value of the abscissa, and Y\(_{min}\) is the minimum value of the ordinate among the coordinates of the four vertices.

When calculating adjacent positions, it may lead to some inaccurate matches. To avoid such situations, the paper detects the minimum parent rectangle of the cropped raster image and calculates the center point of the rectangle. Using this method, the feature point information for each parent rectangle can be calculated. The calculation formula is as follows:

\begin{equation}
	\begin{aligned}
		&\left\{
		\begin{aligned}
			&f(A,B)=(Bx-Ax)(py-Ay)-(px-Ax)(By-Ay) \\
			&f(p_2,p_3)\times f(p_4,p_3)\geq0 \\
			&f(p_2,p_3)\times f(p_4,p_1)\geq0
		\end{aligned}  
		\right.
	\end{aligned}
\end{equation}

In the formula, A and B represent two adjacent vertices of the rectangle, while x and y represent the abscissa and ordinate of any point inside the rectangle. p\(_1\) to p\(_4\) are the four vertices of the rectangle, and p is a feature point. If equation (1) is satisfied, it indicates that the feature point is located inside the rectangle.

\subsection{Image Rasterization}
Image rasterization is the process of dividing an image into several pixel units and representing each pixel unit as a fixed-size square. This process usually involves aspects such as image segmentation, feature extraction, pattern recognition, etc., and is an important research direction in the field of digital image processing. In Python, the Pillow library can be used to implement image rasterization operations, including opening the image, resizing, converting the image, obtaining pixel color values, and converting them to binary (0/1). Image rasterization is widely used in computer vision, machine learning, and other fields, such as object detection, image classification, image recognition, etc.

The rasterization of indoor maps involves transforming a two-dimensional map into pixel grid data. The basic steps of rasterization methods include binarization, grayscale conversion, feature extraction, and image segmentation. Binarization converts the image into black and white, with pixel values of only 0 and 1. Grayscale conversion transforms the image into a grayscale image with pixel values ranging from 0 to 255. Feature extraction extracts the feature information of the indoor map, such as walls, doors, etc., based on the grayscale image. Image segmentation divides the indoor map into multiple connected regions for subsequent processing and storage.

The specific steps of image rasterization are as follows:
\begin{enumerate}[label=(\arabic*)]
	\item Open the map image,
	\item Resize the map image to the desired size,
	\item Convert the map image to grayscale,
	\item Obtain the color value of each pixel and convert it to binary,
	\item Save or use the binary pixel matrix.
\end{enumerate}

The example of map image rasterization is shown in Figures \ref{fig2} and \ref{fig3}:
\begin{figure}[ht]
	\centering
	\includegraphics[width=.8\linewidth]{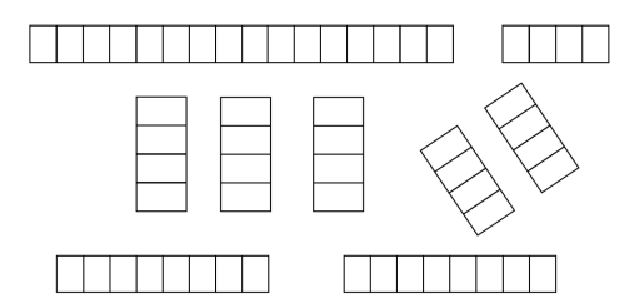}
	\caption{Map grayscale processing} 
	\label{fig2}
\end{figure}
\begin{figure}[ht]
	\centering
	\includegraphics[width=.8\linewidth]{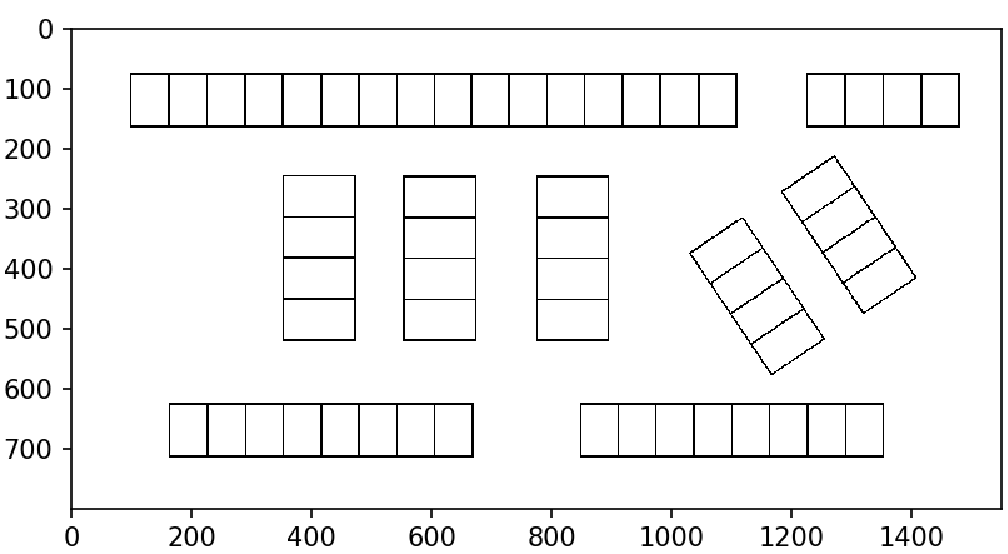}
	\caption{The map images were reconstructed using a binary matrix} 
	\label{fig3}
\end{figure} 

\section{Vectorization and Categorical Storage of Garage Map Raster Data}\label{sec4}
\subsection{Raster Data Vectorization}
Raster data vectorization is the process of converting raster data into vector data. In GIS, raster data consists of grid-like data composed of pixels, while vector data consists of spatial data composed of basic elements such as points, lines, and polygons. Converting raster data to vector data allows for more convenient spatial analysis and modeling, as vector data offers better precision and visualization effects.

The main steps of raster data vectorization include:
\begin{enumerate}[label=(\arabic*)]
	\item Pixel Classification: Each pixel in raster data is classified based on its attribute values, such as classifying it as a parking space, obstacle, path, etc.
	\item Edge Detection \cite{li2021measurement}: For each classified pixel, compare it with neighboring pixels to detect boundary lines.
	\item Boundary Refinement: Refine and repair the detected boundary lines to ensure they are continuous and smooth.
	\item Vectorization: Convert the boundary lines into vector basic elements such as points, lines, and polygons to generate vector data.
	\item Data Editing: Edit and repair the generated vector data to ensure its accuracy and completeness.
\end{enumerate}
\subsection{Map Vector Data Processing}
Image processing techniques are used to process the garage map. The edge detection algorithm is applied to extract the edges of the garage map, followed by using morphological transformation algorithms to binarize the map. Subsequently, the connected component algorithm is employed to separately extract elements such as parking spaces, pathways, and obstacles, transforming them into labeled vector data.

To extract pathways in the map, the connected component algorithm \cite{lv2017fast} can be used. The basic idea of the algorithm is to traverse the two-dimensional matrix, find all connected 0s, and form independent regions. Then, contour extraction is performed on each region, obtaining the vector representation of the region, which is then converted into vector data and stored in a database.

To extract parking spaces in the map, the polygon fitting algorithm \cite{mao2005polygon} can be used. Detecting rectangular parking spaces using the polygon fitting algorithm involves the following steps:
\begin{enumerate}[label=(\arabic*)]
	\item For each connected region (pathway), use contour detection algorithm to obtain its set of contour points.
	\item For the contour point set of each connected region, use the polygon fitting algorithm to fit a convex hull polygon.
	\item For the fitted convex hull polygon, calculate its perimeter and area.
	\item For convex hull polygons that meet certain area and perimeter restrictions, determine whether they are rectangular, i.e., whether the four corner points are approximately right angles.
	\item For convex hull polygons meeting the rectangular conditions, extract their four corner coordinates to obtain rectangular parking spaces, and use the grid index method to extract the coordinates of parking spaces and obstacles.
\end{enumerate}

\begin{figure}[ht] 
	\centering
	\includegraphics[width=.8\linewidth]{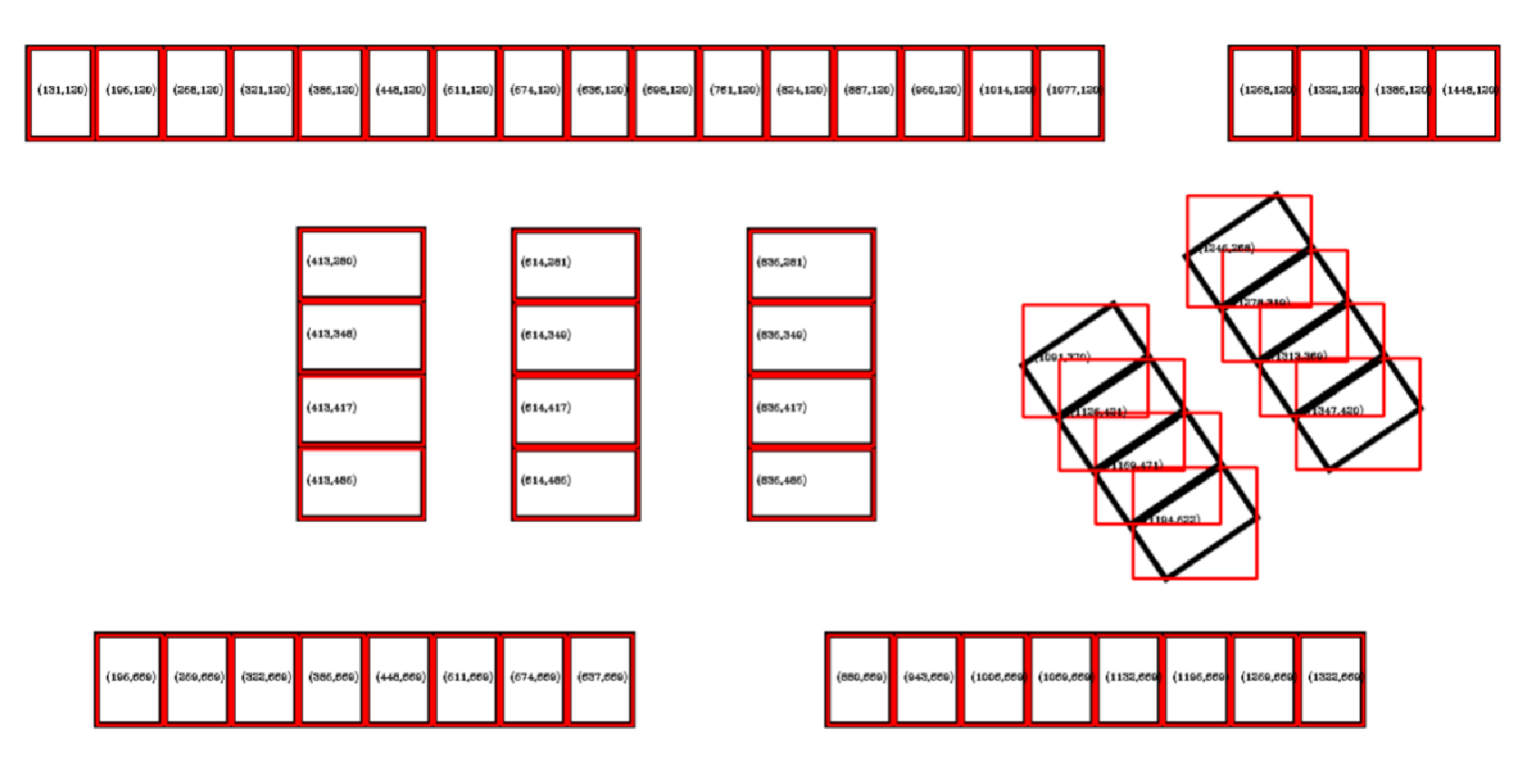}
	\caption{The location coordinates of the parking space were extracted by shape detection} 
	\label{fig4}
\end{figure}

The map's vector data processing is shown in Figure \ref{fig4}. After obtaining the parking space matrix, the system extracts the corresponding coordinates for each parking space. First, based on the position information of the parking space matrix, the system determines the geometric position of each parking space on the map. Then, through scanning and analyzing the map, it extracts the specific coordinate information for each parking space. This coordinate information can include the latitude and longitude of the parking space or the pixel coordinates relative to the map. The extracted parking space coordinates are stored in the system's database for subsequent parking navigation and path planning. By extracting parking space coordinates, the system can accurately locate each parking space, providing precise parking navigation and path planning services for vehicle owners.
\subsection{Data Classification and Storage}
This paper uses the grid index method to classify and store elements such as parking spaces, pathways, and obstacles based on their coordinate positions. Firstly, the garage map is divided into several grids, with each grid size equal to that of a parking space. Then, parking spaces, pathways, and obstacles are placed into their respective grids based on their coordinate positions, forming a two-dimensional array. This greatly improves the efficiency of querying parking spaces, pathways, and obstacles since only the relevant grid needs to be queried.

A SQL database model is designed, including tables for parking spaces, pathways, and obstacles. The parking space table includes information such as parking space number, coordinate position, and parking space type. The pathway table includes information such as pathway number, starting point coordinates, and ending point coordinates. The obstacle table includes information such as obstacle number, coordinate position, and obstacle type. Parking space types and obstacle types can be determined based on features such as color or shape, for example, red parking spaces are no-parking zones, rectangular parking spaces are for small vehicles, and circular parking spaces are for large vehicles. Model construction examples are shown in Tables \ref{tab1}, \ref{tab2} and \ref{tab3}.

\begin{table}[h]
	\caption{Parking table}\label{tab1}%
	\begin{tabular}{@{}llll@{}}
		\toprule
		Attribute Name&Data Type&Primary Key&Description\\
		\midrule
		id&int&YES&Parking Space ID\\
		x\_coordinate&double&-&Parking Space X Coordinate\\
		y\_coordinate&double&-&Parking Space Y Coordinate\\
		space\_type&char&-&Parking Space Type\\
		\botrule
	\end{tabular}
\end{table}
\begin{table}[h]
	\caption{Routing table}\label{tab2}%
	\begin{tabular}{@{}llll@{}}
		\toprule
		Attribute Name&Data Type&Primary Key&Description\\
		\midrule
		id&int&YES&Parking ID\\
		start\_x&double&-&Path Start X Coordinate\\
		start\_y&double&-&Path Start Y Coordinate\\
		end\_x&double&-&Path Start x Coordinate\\
		end\_y&double&-&Path Start y Coordinate\\
		\botrule
	\end{tabular}
\end{table}
\begin{table}[h]
	\caption{Obstruct table}\label{tab3}%
	\begin{tabular}{@{}llll@{}}
		\toprule
		Attribute Name&Data Type&Primary Key&Description\\
		\midrule
		id&int&YES&Parking ID\\
		x\_coordinate&double&-&Obstacle X Coordinate\\
		y\_coordinate&double&-&Obstacle Y Coordinate\\
		obstacle\_type&char&-&Obstacle Type\\
		\botrule
	\end{tabular}
\end{table}

\section{Extraction of Vector Data and its Conversion to Raster Data}\label{sec5}
Vector data extraction and conversion to raster data can be achieved using interpolation algorithms. Interpolation is a method of estimating unknown position data by leveraging the values of known data points. It involves inferring spatial relationships between known data points to compute the values of unknown positions. When transforming vector data into raster data, interpolation algorithms can be employed to calculate the values for each raster grid.

One common interpolation algorithm is Inverse Distance Weighting (IDW), which calculates the value of unknown positions based on the weighted distances. When converting vector data to raster data, each vector data point can be treated as a known data point. The values for each raster grid are then computed based on the coordinates and attribute values of these known data points. Specifically, for a raster grid position, the distances to all known data points are calculated. The weights are determined based on these distances and the attribute values of the known data points, and the raster grid value is subsequently computed using these weights.
\subsection{Vector Point Rasterization}
\begin{figure}[ht] 
	\centering
	\includegraphics[width=.6\linewidth]{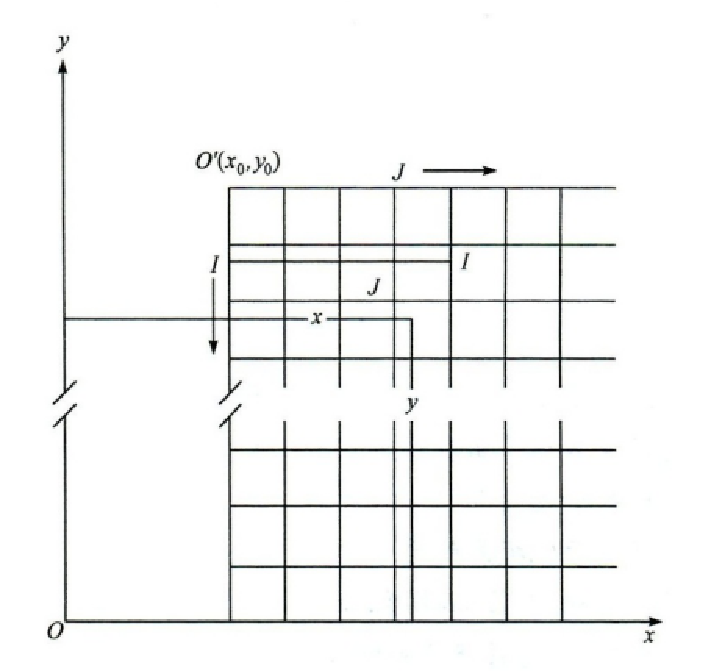}
	\caption{The vector data points were converted to raster data} 
	\label{fig5}
\end{figure}

Vector point rasterization: Essentially involves converting the coordinates (x,y) of points to row and column numbers in a raster. The coordinate origin of the raster is located at the top-left corner, where the row number I increases downward, and the column number J increases to the left. Vector point rasterization is illustrated in Figure \ref{fig5}.

In the figure, (x\(_0\),y\(_0\))represents the origin of the raster, while d\(_x\) and d\(_y\) respectively represent the length and width of the raster, i.e., its size. During the conversion of vector data to raster data, raster data can be expressed using a two-dimensional array.

\subsection{Vector Line Rasterization}
\begin{figure}[ht] 
	\centering
	\includegraphics[width=.6\linewidth]{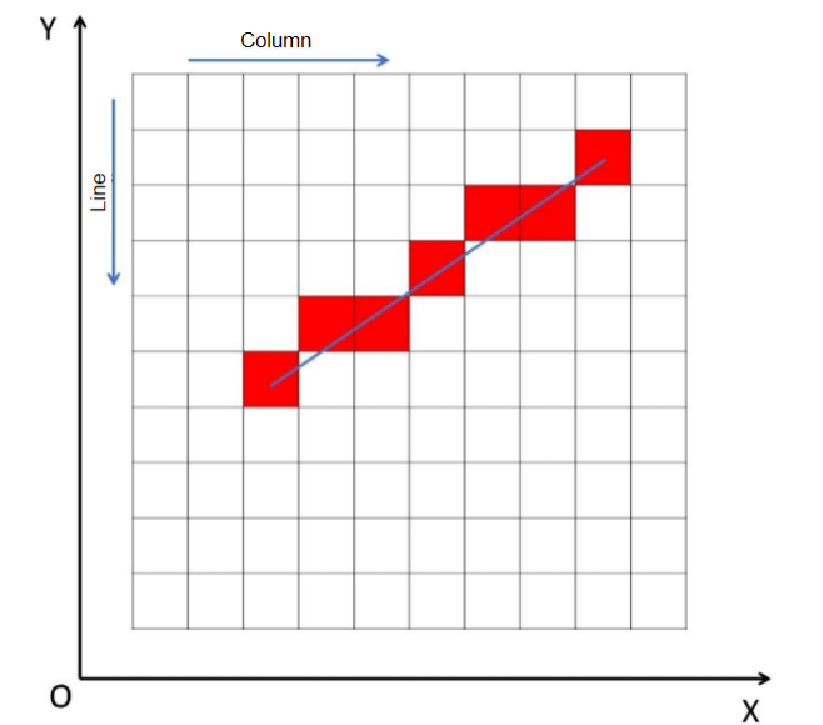}
	\caption{Eight direction grid} 
	\label{fig6}
\end{figure}

Eight-Direction Rasterization:It is a straightforward method based on the inclinations of lines.It ensures that only one pixel is marked as black in each row or column. This method decomposes a line into eight fixed directions (horizontal, vertical, and diagonal) and determines the pixel positions on each direction based on the inclination of the line. Figure \ref{fig6} illustrates an example of eight-direction rasterization. Using this algorithm, lines can be effectively converted into discrete pixel representations, facilitating subsequent processing and analysis.

\begin{figure}[ht] 
	\centering
	\includegraphics[width=.6\linewidth]{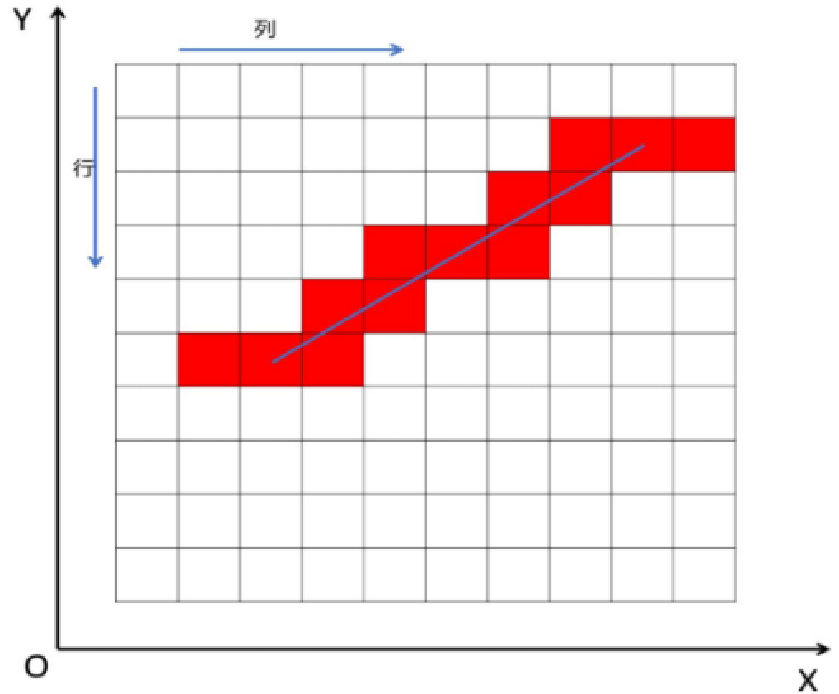}
	\caption{Full path raster} 
	\label{fig7}
\end{figure}

Full Path Rasterization: This is essentially a "striping" method, where the raster is divided into rows or columns, and all elements intersecting with the vector line on each strip are filled. As shown in Figure \ref{fig7}.
\subsection{Two-Dimensional Polygon Sampling}
Grid sampling is a common method for two-dimensional polygon sampling, enabling the transformation of raster data into regularly sampled grid points. The basic steps for creating grid sampling for maps are as follows:

Create Grid Sampling: Based on the sampling interval and the bounding box of polygon features, create regularly spaced grid points within each polygon's range. The following steps can be used for grid sampling:

Determine the sampling interval: Based on the needs and precision of the map, determine the spacing between grid points, such as the horizontal and vertical intervals within each polygon.

Iterate through polygon features: For each polygon feature, create grid points within the bounding box of the feature at the specified sampling interval. Use two nested loops to iterate through the coordinates within the bounding box and determine the position of grid points based on the sampling interval.

\begin{figure}[ht] 
	\centering
	\includegraphics[width=.9\linewidth]{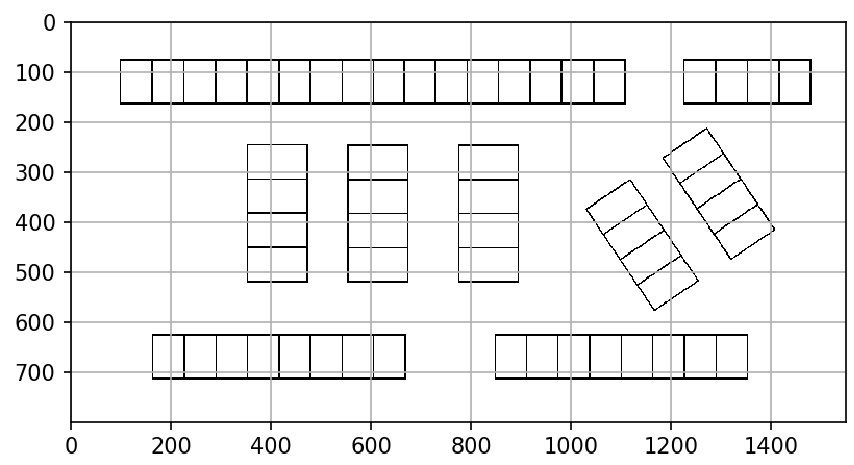}
	\caption{Example plots of the grid sampling} 
	\label{fig8}
\end{figure}

Determine grid point positions: During the iteration, use the current coordinates as the position of the grid points, and add each point to the sampling point collection. An example of creating grid coordinate sampling is shown in Figure \ref{fig8}.

\section{Establishment of Map Models and Navigation Testing}\label{sec6}
Navigational testing of the map stored in the database through vector rasterization can be conducted following these steps:

Extract map information: Extract map information from the database stored in vector rasterized format and convert it into the required format for navigation. For example, road information can be transformed into a graph of nodes and edges, establishing a road network.

Set start and end points: Set start and end points on the road network graph, and determine the navigation algorithm.

Navigation algorithm: Utilize the navigation algorithm based on the set start and end points to calculate the shortest or optimal path.

Visualize navigation results: Visualize the navigation results to provide users with a clear view of the navigation path, including navigation instructions. For instance, display the navigation path, turning directions, distance, time, etc., on the map.

Navigation testing: Conduct navigation testing to examine the accuracy and reliability of the navigation results.

The navigation test is illustrated in Figures \ref{fig9} and \ref{fig10}:
\begin{figure}[ht] 
	\centering
	\includegraphics[width=.8\linewidth]{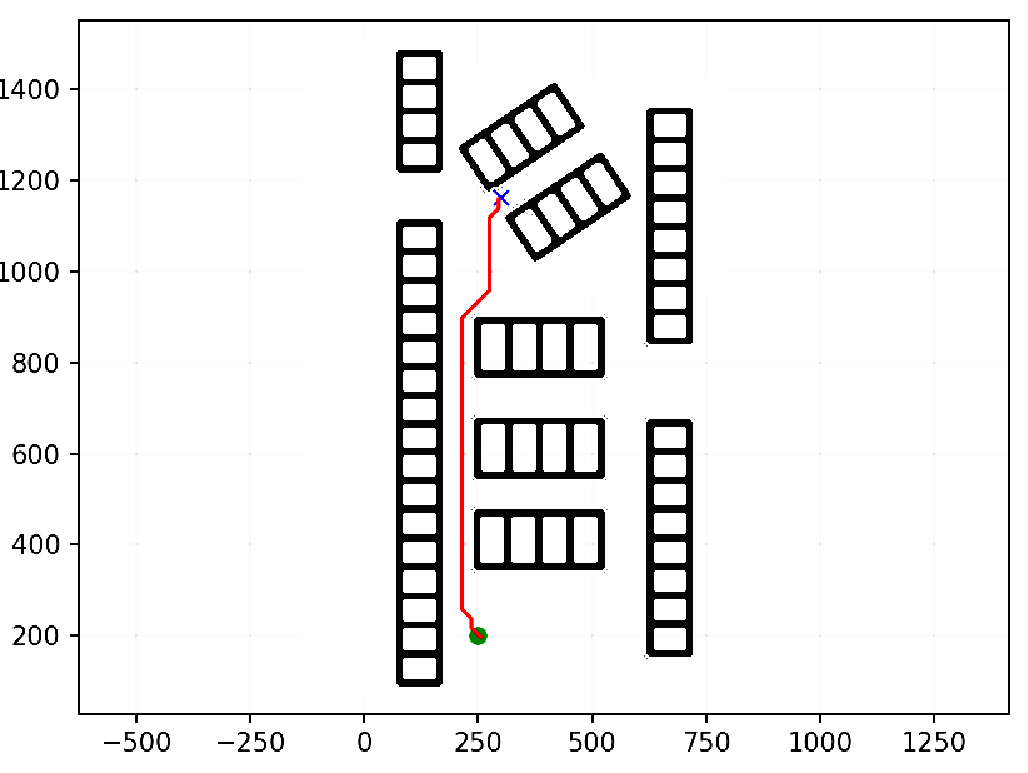}
	\caption{Vehicle navigation example diagram1} 
	\label{fig9}
\end{figure} 

\begin{figure}[ht] 
	\centering
	\includegraphics[width=.8\linewidth]{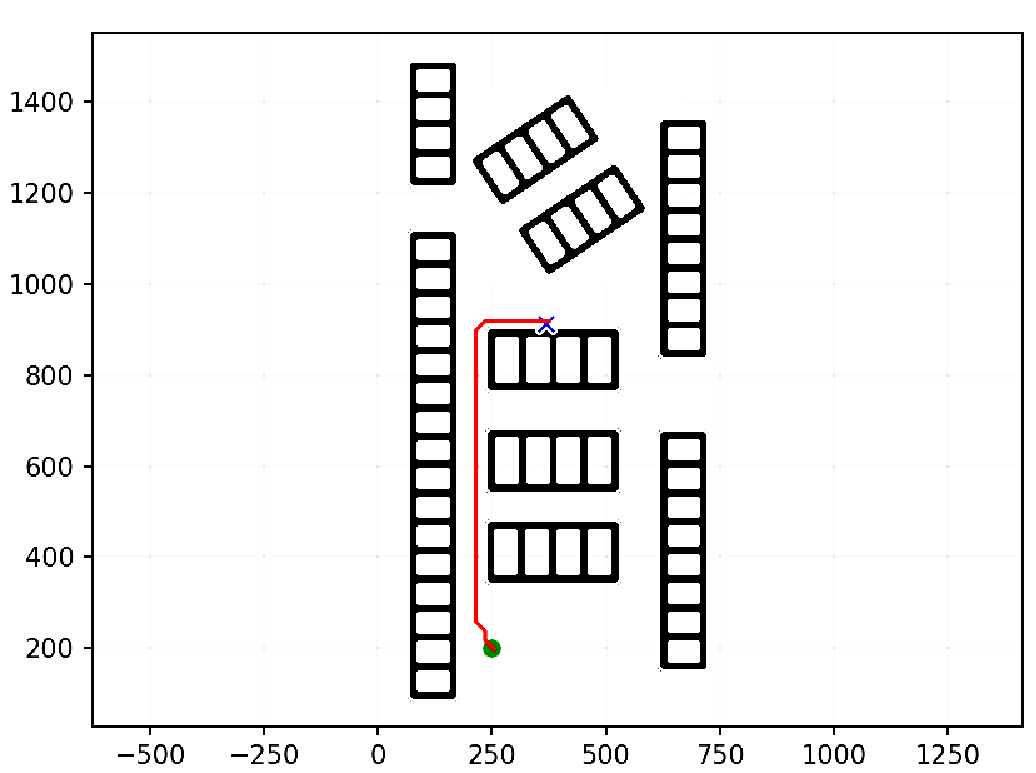}
	\caption{Vehicle navigation example diagram2} 
	\label{fig10}
\end{figure} 

\section{Conclusion}\label{sec7}
In this study, a method for the structural mapping of a large underground parking garage was proposed, involving image processing and vector data conversion to achieve systematic map modeling. Here is the summary of the method:

Image feature retrieval: Correction of raster images and latitude-longitude template images is performed first. Subsequently, image feature retrieval is employed to calculate feature points of the parent matrix, and matching is conducted at last. Through the matched feature point correspondences, GIS features on the raster image and their Cartesian coordinates are determined.

Affine transformation model solution: Utilizing the matching results, the data for the affine transformation model is computed. By establishing the affine transformation matrix, pixel coordinates in the raster image can be mapped to actual geographic coordinates.

Raster image vectorization: Based on matching and affine transformation results, the raster image is converted into vector data. Pixel points in the raster image are connected to form line segments or polygonal elements, creating vector polygon features. Each feature represents structures such as entrances, exits, or parking spaces in the garage.

Vector data classification and storage: Grid indexing is employed to classify and store elements like parking spaces, pathways, and obstacles based on their coordinate positions.

Coordinate transformation: Conversion of coordinate information from the raster image to coordinate values compatible with the map coordinate system. This allows for map projection and spatial analysis of the vectorized data.


\bibliography{ref}                        

\end{document}